# A Split-Merge Framework for Comparing Clusterings

Qiaoliang Xiang[1]                                                  QIAOLIANGXIANG@GMAIL.COM
Qi Mao[1]                                                                  QMAO1@NTU.EDU.SG
Kian Ming A. Chai[2]                                                CKIANMIN@DSO.ORG.SG
Hai Leong Chieu[2]                                                  CHAILEON@DSO.ORG.SG
Ivor Wai-Hung Tsang[1]                                          IVORTSANG@NTU.EDU.SG
Zhendong Zhao[3]                                                    ZHENDONG.ZHAO@MQ.EDU.AU

[1] School of Computer Engineering, Nanyang Technological University, Singapore
[2] DSO National Laboratories, Singapore
[3] Department of Computing, Macquarie University, Australia

## Abstract

Clustering evaluation measures are frequently used to evaluate the performance of algorithms. However, most measures are not properly normalized and ignore some information in the inherent structure of clusterings. We model the relation between two clusterings as a bipartite graph and propose a general component-based decomposition formula based on the components of the graph. Most existing measures are examples of this formula. In order to satisfy consistency in the component, we further propose a split-merge framework for comparing clusterings of different data sets. Our framework gives measures that are conditionally normalized, and it can make use of data point information, such as feature vectors and pairwise distances. We use an entropy-based instance of the framework and a coreference resolution data set to demonstrate empirically the utility of our framework over other measures.

## 1. Introduction

Hard partitional clustering groups data points into a set of disjoint clusters. There are three types of measures which could be used to evaluate a clustering: 1) an external measure that compares the clustering to a given true clustering; 2) an internal measure which only utilizes the information of feature vectors of data points; 3) a hybrid measure which takes both information into account. External measures are preferred because they better reflect human evaluation (Strehl & Ghosh, 2003).

Clustering evaluation measures are commonly used to compare the performance of various algorithms, so they should be able to compare clusterings of different data sets. Unnormalized and asymmetric measures are inappropriate for comparing clusterings across data sets (Vinh et al., 2010; Wagner & Wagner, 2007). Therefore, measures should be normalized properly and be independent of some inherent structures of two clusterings (Meilă, 2007). When a measure is used for comparing all possible clusterings with the true clustering, it is preferable that the similarity scores should be in the closed interval $[0, 1]$ (Luo et al., 2009). Besides, a measure should not depend on some parameters such as the number of data points (Meilă, 2007).

Existing external measures can be grouped into three categories: pair counting, set matching and information theoretic. Pair counting measures are based on counting the pairs of points for which two clusterings agree or disagree. They are sensitive to parameters, such as the size of a cluster, the number of clusters, and the number of data points (Wagner & Wagner, 2007). Set matching measures find a maximum matching between two clusterings. They make no assumption on how clusterings are generated, but they ignore those unmatched clusters (Meilă, 2007). Moreover, since the matching degree between two clusterings is always positive, such measures are not normalized. Information theoretic measures do not suffer from the problems of pair counting and set matching measures. They have been analyzed extensively and systematically in recent years (Meilă, 2007; Vinh et al., 2010). Some measures tend to give high scores in practice, so adjusted measures, such as adjusted Rand index (Hubert & Arabie, 1985) and adjusted mutual information (Vinh et al., 2010) are proposed to address this issue, but they are not normalized because they may be negative (Meilă, 2007).

In this paper, instead of focusing on designing a new clustering measure, we propose a *split-merge framework* that can be tailored to different applications (Guyon et al., 2009). The framework models two clusterings as a bipartite graph which is decomposed into connected components, and each component is further decomposed into subcomponents. Pairs of related subcomponents are then taken into consideration in designing a clustering similarity measure within the framework. The contributions of this paper are listed below.

- We propose a general component-based decomposition formula based on the components of the bipartite graph. We find that most existing measures are special cases of that formula.
- The framework can compare clusterings across data sets. It is join-weighted decomposable on components (Property 4.1), consistent between a component and its subcomponents (Property 4.4), and conditionally normalized (Property 5.1). Moreover, it satisfies many properties of the variation of information (Meilă, 2007).
- The framework is flexible and easy to use. It can be instantiated by providing a measure to score a subcomponent, which contains a cluster and a partition of the cluster. It is relatively easier to design such a measure. For example, one can either make use of an existing measure or additional information of data points, such as feature vectors or pairwise distances.

The rest of the paper is organized as follows. Section 2 introduces some relevant concepts and notations. Some representative measures are discussed in Section 3. We present the split-merge framework in Section 4 and compare it with other measures in Section 5. Experimental results are given in Section 6, and Section 7 concludes this work.

## 2. Preliminaries

Let $D = \{1, 2, \ldots, n\}$ be a set of $n$ data points, and let the feature vector of the $i$-th point be denoted by $f_i$. A clustering is a set of clusters, and a cluster is a set of points. Let $\Omega$ be the set of all clusterings, $\mathbb{L} \in \Omega$ be the true clustering and $\mathbb{C} \in \Omega$ be the predicted clustering. $L$ (resp. $C$) denotes any cluster of $\mathbb{L}$ (resp. $\mathbb{C}$). Denote an empty clustering by $\varnothing$ and an empty cluster by $\emptyset$. A cluster is a *singleton* if it contains only one data point. The term $\binom{n}{2} = n(n-1)/2$ is the number of pairwise links between $n$ points. The entropy of a clustering $\mathbb{L}$ is $H(\mathbb{L}) = -\sum_{L \in \mathbb{L}} \frac{|L|}{n} \log \frac{|L|}{n}$, while the joint entropy between two clusterings $\mathbb{L}$ and $\mathbb{C}$ is $H(\mathbb{L}, \mathbb{C}) = -\sum_{L \in \mathbb{L}} \sum_{C \in \mathbb{C}} \frac{|L \cap C|}{n} \log \frac{|L \cap C|}{n}$. The amount of information shared between $\mathbb{L}$ and $\mathbb{C}$ is $I(\mathbb{L}, \mathbb{C}) = H(\mathbb{L}) + H(\mathbb{C}) - H(\mathbb{L}, \mathbb{C})$. The conditional entropy of $\mathbb{C}$ given $\mathbb{L}$ is $H(\mathbb{C}|\mathbb{L}) = H(\mathbb{C}) - I(\mathbb{C}, \mathbb{L})$ (Cover & Thomas, 1991).

We introduce some relevant concepts from lattice theory (Grätzer, 2011). *Top* $\top$ is the clustering that groups all the points into a cluster, and *bottom* $\bot$ is the clustering that treats each point as a singleton (Grätzer, 2011). $\mathbb{C}$ *refines* $\mathbb{L}$ if $\mathbb{C}$ can be obtained by only splitting one or more clusters of $\mathbb{L}$. The *meet* $\mathbb{M} = \{L \cap C \mid L \in \mathbb{L}, C \in \mathbb{C}, L \cap C \neq \emptyset\}$ is the clustering that contains all nonempty intersections of every cluster from $\mathbb{L}$ with every cluster from $\mathbb{C}$. The *join* $\mathbb{J}$ is the clustering with the greatest number of clusters that is refined by both $\mathbb{L}$ and $\mathbb{C}$. $J$ denotes any cluster of $\mathbb{J}$. Note that both $\mathbb{M}$ and $\mathbb{J}$ are partitions of $D$.

## 3. Related Work

In this paper, we focus on studying symmetric similarity measures. For a distance measure, we study its counterpart similarity measure by subtracting it from one. Meilă (2007); Wagner & Wagner (2007); Vinh et al. (2010) summarized and compared a large number of measures that have been proposed in the literature, and a few representative measures are discussed as follows.

### 3.1. Pair Counting Measures

**Rand Index** A link is positive if the two points are within the same cluster, otherwise it is negative. There are $P(\mathbb{L}, \mathbb{C}) = \sum_{L \in \mathbb{L}} \sum_{C \in \mathbb{C}} \binom{|L \cap C|}{2}$ positive links and $\binom{n}{2} - P(\mathbb{L}, \mathbb{L}) - P(\mathbb{C}, \mathbb{C}) + P(\mathbb{L}, \mathbb{C})$ negative links common to two clusterings. Rand index is the fraction $R(\mathbb{L}, \mathbb{C}) = (\binom{n}{2} - P(\mathbb{L}, \mathbb{L}) - P(\mathbb{C}, \mathbb{C}) + 2P(\mathbb{L}, \mathbb{C}))/\binom{n}{2}$ of links common to two clusterings (Rand, 1971). It is large when there are many clusters (Wagner & Wagner, 2007).

### 3.2. Set Matching Measures

**Van Dongen Criterion** In order to transform $\mathbb{L}$ to $\mathbb{C}$, $2n - \sum_{L \in \mathbb{L}} \max_{C \in \mathbb{C}} |L \cap C| - \sum_{C \in \mathbb{C}} \max_{L \in \mathbb{L}} |L \cap C|$ point moves are required (Dongen, 2000). This metric can be constrained to a right-open interval $[0, 1)$ by dividing by $2n$ (Meilă, 2007). The similarity counterpart is $N(\mathbb{L}, \mathbb{C}) = \frac{1}{2} \sum_{L \in \mathbb{L}} \max_{C \in \mathbb{C}} \frac{|L \cap C|}{n} + \frac{1}{2} \sum_{C \in \mathbb{C}} \max_{L \in \mathbb{L}} \frac{|L \cap C|}{n}$. It is not normalized because its lower bound is nonzero.

**Classification Accuracy** By considering clustering as a classification task, classification accuracy computes the fraction of points that are correctly classified (Meilă & Heckerman, 2001). Finding the best mapping between two clusterings is equivalent to solving a maximum weighted bipartite matching problem (Meilă, 2005). The classification accuracy is $A(\mathbb{L}, \mathbb{C}) = \max_W \sum_L \sum_C W(L, C) \frac{|L \cap C|}{n}$ subject to $\forall L \forall C, W(L, C) \in \{0, 1\}$; $\forall L, \sum_C W(L, C) = 1$; and $\forall C, \sum_L W(L, C) = 1$, and $C$ (resp. $L$) ranges over $\mathbb{C}$ (resp. $\mathbb{L}$). Its lower bound is $1/n$ instead of zero.

## 3.3. Information Theoretic Measures

**Normalized Mutual Information** Vinh et al. (2010) advocated $NMI(\mathbb{L}, \mathbb{C}) = I(\mathbb{L}, \mathbb{C}) / \max\{H(\mathbb{L}), H(\mathbb{C})\}$ after comparing several normalized variants of the mutual information. The issue with $NMI(\mathbb{L}, \mathbb{C})$ is that $I(\mathbb{L}, \mathbb{C})$ indicates the degree of statistical dependency between two clusterings, and this is not always consistent with their similarity. For example, $I(\top, \mathbb{C}) = 0$ means there is no dependency between $\top$ and any $\mathbb{C} \in \Omega$, but the actual similarity depends on the closeness of $\mathbb{C}$ to $\top$.

**Normalized Variation of Information** The variation of information $VI(\mathbb{L}, \mathbb{C}) = H(\mathbb{C}|\mathbb{L}) + H(\mathbb{L}|\mathbb{C})$ is the change in the amount of information when transforming $\mathbb{L}$ into $\mathbb{C}$ (Meilă, 2007). Although $VI(\mathbb{L}, \mathbb{C})$ has certain desired properties, it is unnormalized. This can be rectified through dividing by the upper bound $\log n$ (Meilă, 2007). Subtracting it from unity gives the similarity measure $V(\mathbb{L}, \mathbb{C}) = 1 - VI(\mathbb{L}, \mathbb{C}) / \log n$. However, this is not suitable for comparing clusterings across data sets due to the dependence on $n$ (Meilă, 2007).

When both $\mathbb{L}$ and $\mathbb{C}$ have at most $k \leq \sqrt{n}$ clusters, $VI(\mathbb{L}, \mathbb{C})$ is upper bounded by $2 \log k$ (Meilă, 2007). Thus, $K(\mathbb{L}, \mathbb{C}) = 1 - VI(\mathbb{L}, \mathbb{C}) / \log k^2$ is an alternative similarity measure that is suitable for comparing clusterings across data sets, but applying this in practice requires knowing $k$ in advance (Meilă, 2007).

# 4. The Split-Merge Framework

In Section 4.1, we model the relation between two clusterings as a bipartite graph that can be decomposed into connected components. A component is further decomposed into split and merge subcomponents in Section 4.2. The split and merge subcomponents can be combined into a derivation graph $\mathcal{D}$ that transforms $\mathbb{L}$ into $\mathbb{C}$. In Section 4.3, we capture the essence of $\mathcal{D}$ by pairing split subcomponents with merge subcomponents. The similarity of each pair, called a subcomponent pair, is discussed in Section 4.4. It is also shown that the precise definition for our split-merge framework is given. Two instances of our framework are given in Section 4.5.

## 4.1. Connected Components of a Bipartite Graph

A bipartite graph governs the relation between $\mathbb{L}$ and $\mathbb{C}$:

**Definition 4.1 (Bipartite Graph).** Given clusterings $\mathbb{L}$ and $\mathbb{C}$, a directed bipartite graph $\mathcal{G} = (\mathbb{L}, \mathbb{C}, \mathbb{E})$ is constructed: $\mathbb{L}$ and $\mathbb{C}$ are the two disjoint sets of vertices, and $\mathbb{E} = \{\langle L, C \rangle \mid L \in \mathbb{L}, C \in \mathbb{C}, L \cap C \neq \emptyset\}$ is the set of directed edges from $L$ to $C$.

**Definition 4.2 (Induced Clustering).** A clustering $\mathbb{C}$ gives an induced clustering $\mathbb{C}_A = \{A \cap C \mid C \in \mathbb{C}, A \cap C \neq \emptyset\}$

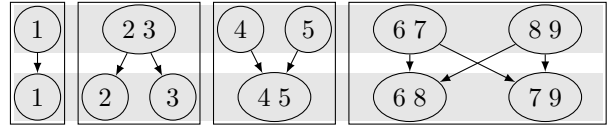

*Figure 1.* The bipartite graph of two clusterings. A cluster is represented by a circle or an ellipse. Clusters on the same rectangle filled in gray belong to the same clustering. The top row is $\mathbb{L}$ while the bottom row is $\mathbb{C}$. Clusters in the same hollow rectangle belong the same component.

when acted upon by a cluster $A$ (Meilă, 2007).

**Definition 4.3 (Induced Subgraph).** $\mathcal{G}_A = (\mathbb{L}_A, \mathbb{C}_A, \mathbb{E}_A)$ is a subgraph induced on $\mathcal{G} = (\mathbb{L}, \mathbb{C}, \mathbb{E})$ by a cluster $A$, where $\mathbb{L}_A$ and $\mathbb{C}_A$ are induced clusterings, and $\mathbb{E}_A = \{\langle L, C \rangle \mid L \in \mathbb{L}_A, C \in \mathbb{C}_A, L \cap C \neq \emptyset\}$ is the set of the remaining edges on the induced clusterings.

**Proposition 4.1 (Component).** $\{\mathcal{G}_J \mid J \in \mathbb{J}\}$ *is the set of connected components of graph* $\mathcal{G}$.

*Proof.* Construct a graph $\mathcal{G}'$ by letting all directed edges of graph $\mathcal{G}$ be undirected. Constructing a clustering that is refined by both $\mathbb{L}$ and $\mathbb{C}$ requires that all reachable clusters of $\mathcal{G}'$ should be grouped together. Many such clusterings can be obtained, and the join $\mathbb{J}$ is the one with the largest number of clusters. So all non-reachable clusters of $\mathcal{G}'$ should not be grouped together. This process is equivalent to finding the connected components of graph $\mathcal{G}$. □

Throughout the paper, *component* means *weakly connected component*. Figure 1 gives a bipartite graph and its components. Denote a general similarity measure by $S(\mathbb{L}, \mathbb{C})$. The similarity score of $\mathcal{G}_J$ is defined as the similarity $S(\mathbb{L}_J, \mathbb{C}_J)$ between $\mathbb{L}_J$ and $\mathbb{C}_J$. Most measures on clusterings can be expressed as the component-based decomposition formula

$$S(\mathbb{L}, \mathbb{C}) = \sum_{J \in \mathbb{J}} w(J, n) S(\mathbb{L}_J, \mathbb{C}_J) + b(\mathbb{J}, n), \quad (1)$$

where $w(J, n)$ is the weight for component $\mathcal{G}_J$, and $b(\mathbb{J}, n)$ is independent of the component scores. Table 1 lists some measures and their decompositions, which are shown in Section 2 of the supplementary material.

In this paper, we propose a restriction on the class of $S(\mathbb{L}, \mathbb{C})$ given in (1). We opine that $S(\mathbb{L}, \mathbb{C})$ should simply be the weighted average of the similarity scores of all components $\{\mathcal{G}_J | J \in \mathbb{J}\}$. That is $b(\mathbb{J}, n) = 0$. Moreover, the importance of component $\mathcal{G}_J$ is determined by the importance of all the points within $J$. In the absence of additional information, every point should be treated equally, so $w(J, n)$ is to be proportional to the size of cluster $J$. In summary, we propose the following convex combination.

*Table 1.* The decomposition of various measures; see (1). We were unable to obtain a decomposition for the normalized mutual information (*NMI*). Measures where $w(J, n) = |J|/n$ and $b(\mathbb{J}, n) = 0$ are called join-weighted (Property 4.1).

| Measure(s) | $w(J, n)$ | $b(\mathbb{J}, n)$ |
|---|---|---|
| $R$ | $\frac{|J|(|J|-1)}{n(n-1)}$ | $1 - \sum_{J \in \mathbb{J}} \frac{|J|(|J|-1)}{n(n-1)}$ |
| $I$ | $\frac{|J|}{n}$ | $\log n - \sum_{J \in \mathbb{J}} \frac{|J|}{n} \log |J|$ |
| $V$ | $\frac{|J| \log |J|}{n \log n}$ | $1 - \sum_{J \in \mathbb{J}} \frac{|J| \log |J|}{n \log n}$ |
| $N, A, K, S^*$ | $\frac{|J|}{n}$ | $0$ |

**Property 4.1 (Join-weighted Decomposition).** *A similarity measure $S(\mathbb{L}, \mathbb{C})$ is join-weighted decomposable if*

$$S(\mathbb{L}, \mathbb{C}) = \sum_{J \in \mathbb{J}} \frac{|J|}{n} S(\mathbb{L}_J, \mathbb{C}_J). \qquad (2)$$

If $\mathbb{C}$ refines $\mathbb{L}$, the above property becomes the similarity version of the convex additivity axiom (Meilă, 2007, Axiom A3). Since $\mathbb{J}$ is the least clustering that is refined by both $\mathbb{L}$ and $\mathbb{C}$, $1 - S(\mathbb{L}, \mathbb{C})$ also satisfies the additivity of composition property (Meilă, 2007, Property 8) if $S(\mathbb{L}, \mathbb{C})$ satisfies Property 4.1.

### 4.2. Split and Merge Subcomponents

A component focuses on the clustering-clustering relation. It may be difficult to assign a score to such a relation. Hence, we further break a component into subcomponents, with the focus on the cluster-clustering relations. We define two kinds of subcomponents depending on whether cluster in the relation is from $\mathbb{L}$ or $\mathbb{C}$.

**Definition 4.4 (Split/Merge Graph).** The *split graph* $\langle L, \mathbb{C} \rangle$ is the complete directed bipartite graph from $\{L\}$ to the induced clustering $\mathbb{C}_L$. The *merge graph* $\langle \mathbb{L}, C \rangle$ is the complete directed bipartite graph from the induced clustering $\mathbb{L}_C$ to $\{C\}$.

Conceptually, a split graph maps a cluster $L$ to one or more clusters of $\mathbb{C}$ which overlap with $L$, while a merge graph maps one or more clusters of $\mathbb{L}$ which overlap with $C$ to $C$. For a component $\mathcal{G}_J$, there can be one or more split/merge graphs, which we call its subcomponents.

**Definition 4.5 (Split/Merge Set/Subcomponent).** The *split set* of component $\mathcal{G}_J = (\mathbb{L}_J, \mathbb{C}_J, \mathbb{E}_J)$ is the set $\{\langle L, \mathbb{C}_J \rangle \mid L \in \mathbb{L}_J\}$ of split graphs. The *merge set* of the component is the set $\{\langle C, \mathbb{L}_J \rangle \mid C \in \mathbb{C}_J\}$ of merge graphs. Each element in the split (resp. merge) set is called a split (resp. merge) *subcomponent* of $\mathcal{G}_J$.

Sometimes, we write $\langle L, \mathbb{C} \rangle$ instead of $\langle L, \mathbb{C}_J \rangle$ in the context of a split subcomponent of $\mathcal{G}_J$ since $L \in \mathbb{L}_J$, so $\mathbb{C}_L$ is the same as $(\mathbb{C}_J)_L$. Similarly for the merge subcomponent.

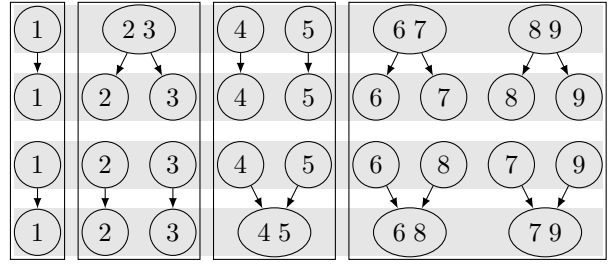

*Figure 2.* Follow on from Figure 1. The top and bottom rows are $\mathbb{L}$ and $\mathbb{C}$ respectively. The middle two clusterings are the same: they are the meet $\mathbb{M}$. Each connected subgraph is a subcomponent within a component. The top graphs are the split graphs while the bottom graphs are the merge graphs. Identifying the two copies of $\mathbb{M}$ together gives the derivation graph $\mathcal{D}$.

**Proposition 4.2.** *The set of sinks in the split set of $\mathcal{G}_J$ is the same as the set of sources in the merge set of $\mathcal{G}_J$. This set is the meet $\mathbb{M}_J$ of $\mathbb{L}_J$ and $\mathbb{C}_J$, and it is identical to the meet $\mathbb{M}$ of $\mathbb{L}$ and $\mathbb{C}$ induced by $J$.*

With the above proposition, we can transform $\mathbb{L}_J$ to $\mathbb{C}_J$ via $\mathbb{M}_J$. The transformation consists of splitting of clusters in $\mathbb{L}_J$ into clusters in $\mathbb{M}_J$ (if necessary), then merging into clusters in $\mathbb{C}_J$ (if necessary). The split and merge mappings are given by the split and merge set of $\mathcal{G}_J$. Figure 2 gives an example. Formally, the transformation follows the derivation graph that combines the split and merge sets.

**Definition 4.6 (Derivation Graph).** The derivation graph $\mathcal{D}$ of $\mathcal{G} = (\mathbb{L}, \mathbb{C}, \mathbb{E})$ is a tripartite graph with parts $\mathbb{L}$, $\mathbb{C}$ and their meet $\mathbb{M}$. The set of subgraphs in $\mathcal{D}$ from $\mathbb{L}$ to $\mathbb{M}$ is the union of the split sets of the components of $\mathcal{G}$ (up to relabeling of the sinks in the union to the clusters in $\mathbb{M}$); and the set of subgraphs from $\mathbb{M}$ to $\mathbb{C}$ is the union of the merge sets (up to relabeling). There is no edge between vertices in $\mathbb{L}$ and vertices in $\mathbb{C}$.

A subcomponent is between a cluster and a clustering, which can be assigned a score more easily than a component. Denote the similarity measure of a split (resp. merge) subcomponent $\langle L, \mathbb{C} \rangle$ (resp. $\langle \mathbb{L}, C \rangle$) by $s(\mathbb{C}|L)$ (resp. $s(\mathbb{L}|C)$). We opine that these measures should include two factors: the number of clusters and the relative size of each cluster (Wagner & Wagner, 2007). Hence, we propose the following property.

**Property 4.2 (Monotonically Decreasing).** *A subcomponent similarity measure is monotonically decreasing if it monotonically decreases as the number of clusters increases or the distribution of cluster sizes gets less skewed.*

The above property becomes the *cluster completeness* (resp. *cluster homogeneity*) constraint (Amigó et al., 2009) when applied to a split (resp. merge) subcomponent. These two constraints are important to clustering measures (Rosenberg & Hirschberg, 2007).

To ensure that $s(\mathbb{C}|L)$ and $s(\mathbb{L}|C)$ are normalized, we propose the following property.

**Property 4.3 (Subcomponent-normalization).** *A split subcomponent similarity measure $s(\mathbb{C}|L)$ is normalized if*

1. *$s(\mathbb{C}|L) = 1$ if and only if $\mathbb{C}_L = \{L\}$;*
2. *$s(\mathbb{C}|L) = 0$ if and only if $\mathbb{C}_L$ has only singletons and $L$ is not a singleton; and*
3. *$s(\mathbb{C}|L) \in (0, 1)$ otherwise.*

*Similarly for the merge subcomponent measure $s(\mathbb{L}|C)$.*

Property 4.2 and Property 4.3 ensure that the similarity measures can be used to score clusterings across data sets.

A split subcomponent $\langle L, \mathbb{C} \rangle$ is a connected bipartite graph by definition. Hence, the similarity measure $S(\mathbb{L}, \mathbb{C})$ is also applicable to it. For consistency, a requirement is that the $S(\{L\}, \mathbb{C}_L)$ evaluates to be the same as $s(\mathbb{C}|L)$. The same must hold for the merge subcomponent.

**Property 4.4 (Subcomponent-consistency).** *A similarity measure $S(\mathbb{L}, \mathbb{C})$ is subcomponent-consistent if $S(\{L\}, \mathbb{C}_L) = s(\mathbb{C}|L)$ and $S(\mathbb{L}_C, \{C\}) = s(\mathbb{L}|C)$.*

### 4.3. Subcomponent Pairs

Within a component $\mathcal{G}_J$, a split (resp. merge) subcomponent may be paired with one or more merge (resp. split) subcomponents in the derivation graph $\mathcal{D}$. If $S(\mathbb{L}_J, \mathbb{C}_J)$ were to be a direct combination of the similarity measures on the subcomponents of $\mathcal{G}_J$, it might give the same value for the different sets of pairings of the subcomponents. This is undesirable. Instead, we propose to base the combination on pairs of split and merge subcomponents.

**Definition 4.7 (Subcomponent Pair).** In a component $\mathcal{G}_J$, a subcomponent pair is the pair $(\langle L, \mathbb{C}_J \rangle, \langle \mathbb{L}_J, C \rangle)$ such that $L \in \mathbb{L}_J$, $C \in \mathbb{C}_J$ and $L \cap C \in \mathbb{M}_J$.

This definition exploits that a split and a merge subcomponent are not disjoint in $\mathcal{D}$ only if a sink in the split subcomponent and a source in the merge subcomponent is the same cluster in $\mathbb{M}_J$; see Proposition 4.2.

**Proposition 4.3.** *For every $M \in \mathbb{M}_J$ there is one and only one subcomponent pair $(\langle L, \mathbb{C}_J \rangle, \langle \mathbb{L}_J, C \rangle)$ such that $L \cap C = M$.*

*Proof.* The existence of the pair is by definition of a subcomponent pair. For uniqueness, suppose there is another $L' \neq L$ where $L' \in \mathbb{L}_J$ and $L' \cap C' = M$ for some $C' \in \mathbb{C}_J$. Then $L' \cap C' = L \cap C$. Since $\mathbb{L}$ is a partition of $D$, $L' \cap L = \emptyset$. So both $L' \cap C'$ and $L \cap C$ must be $\emptyset$ for them to be equal. But $M \neq \emptyset$ by the definition of meet. Hence a contradiction in this case. The other case where $C' \neq C$ gives a contradiction similarly. ☐

With this proposition, we can exactly enumerate all the subcomponent pairs in a component $\mathcal{G}_J$ by enumerating the clusters in $\mathbb{M}_J$. Moreover, $\mathbb{M}_J$ is a partition on $J$. These two properties suggest the following decomposition of $S(\mathbb{L}_J, \mathbb{C}_J)$ in the spirit of Property 4.1:

$$S(\mathbb{L}_J, \mathbb{C}_J)$$
$$= \sum_{M \in \mathbb{M}_J} \frac{|M|}{|J|} \sum_{L \in \mathbb{L}_J, C \in \mathbb{C}_J} \delta(L \cap C, M)\, \sigma(\langle L, \mathbb{C}_J \rangle, \langle \mathbb{L}_J, C \rangle),$$

where $\delta$ is the Kronecker delta, and $\sigma(\cdot, \cdot)$ is the similarity measure on a subcomponent pair $\langle L, \mathbb{C}_J \rangle, \langle \mathbb{L}_J, C \rangle$. We will discuss the choice of $\sigma(\cdot, \cdot)$ in Section 4.4. Using Proposition 4.3 and $|\emptyset| = 0$, the above decomposition can be simplified:

$$S(\mathbb{L}_J, \mathbb{C}_J) = \sum_{L \in \mathbb{L}_J} \sum_{C \in \mathbb{C}_J} \frac{|L \cap C|}{|J|} \sigma(\langle L, \mathbb{C}_J \rangle, \langle \mathbb{L}_J, C \rangle).$$

This can be directly substituted into (2). After subsuming the sum over the join into the sums over the clusterings $\mathbb{L}$ and $\mathbb{C}$, we obtain the following convex combination.

**Property 4.5 (Meet-weighted Decomposition).** *A similarity measure $S(\mathbb{L}, \mathbb{C})$ is a meet-weighted decomposition if*

$$S(\mathbb{L}, \mathbb{C}) = \sum_{L \in \mathbb{L}} \sum_{C \in \mathbb{C}} \frac{|L \cap C|}{n} \sigma(\langle L, \mathbb{C} \rangle, \langle \mathbb{L}, C \rangle). \quad (3)$$

### 4.4. Similarity on Subcomponent Pairs

We now discuss the choice of the similarity measure $\sigma$ on a subcomponent pair. Given a pair $(\langle L, \mathbb{C} \rangle, \langle \mathbb{L}, C \rangle)$, the general idea is to let $\sigma$ be a function of the similarity measures $s(\mathbb{C}|L)$ and $s(\mathbb{L}|C)$ of the subcomponents. We discuss two functions: the product and the arithmetic mean. Our preference is the product because it satisfies subcomponent-consistency (Property 4.4), while the mean does not.

**Product** Our *split-merge framework* defines the meet-weighted decomposed similarity

$$S^*(\mathbb{L}, \mathbb{C}) = \sum_{L \in \mathbb{L}} \sum_{C \in \mathbb{C}} \frac{|L \cap C|}{n} s(\mathbb{C}|L) s(\mathbb{L}|C). \quad (4)$$

where $s(\mathbb{C}|L)$ and $s(\mathbb{L}|C)$ are the subcomponent-normalized similarity measures of the split and merge subcomponents. It is subcomponent-consistent because

$$S^*(\{L\}, \mathbb{C}_L) = \sum_{C \in \mathbb{C}_L} (|C|/|L|) s(\mathbb{C}_L|L) s(\{C\}|C)$$

$$= s(\mathbb{C}_L|L) \left( \sum_{C \in \mathbb{C}_L} (|C|/|L|) \underbrace{s(\{C\}|C)}_{=1} \right) = s(\mathbb{C}|L) \times 1,$$

where $s(\mathbb{C}_L|L) = s(\mathbb{C}|L)$ by definition of the split graph; and similarly for $S^*(\mathbb{L}_C, \{C\})$.

**Arithmetic Mean** One might also use other functions of $s(\mathbb{C}|L)$ and $s(\mathbb{L}|C)$ to define $\sigma$. A natural choice is the arithmetic mean, which combined with (3) gives

$$S'(\mathbb{L}, \mathbb{C}) = \frac{1}{2} \sum_{L \in \mathbb{L}} \frac{|L|}{n} s(\mathbb{C}|L) + \frac{1}{2} \sum_{C \in \mathbb{C}} \frac{|C|}{n} s(\mathbb{L}|C). \quad (5)$$

Some existing measures are instances of $S'$. If we define $s(\mathbb{C}|L) = \max_{C \in \mathbb{C}} \frac{|L \cap C|}{|L|}$ and $s(\mathbb{L}|C) = \max_{L \in \mathbb{L}} \frac{|L \cap C|}{|C|}$, then $S'(\mathbb{L}, \mathbb{C})$ becomes $N(\mathbb{L}, \mathbb{C})$. $S'(\mathbb{L}, \mathbb{C})$ becomes $K(\mathbb{L}, \mathbb{C})$ if we define $s(\mathbb{C}|L) = 1 - H(\mathbb{C}_L)/\log k^2$ and $s(\mathbb{L}|C) = 1 - H(\mathbb{L}_C)/\log k^2$.

The similarity $S'(\mathbb{L}, \mathbb{C})$ directly combines the subcomponent similarities. Hence, it defeats the very purpose of subcomponent pairs. In contrast, the similarity given by (4) cannot be (linearly) decomposed further into subcomponent similarities. Moreover, $S'(\mathbb{L}, \mathbb{C})$ is not subcomponent-consistent: $S'(\{L\}, \mathbb{C}_L) = s(\mathbb{C}|L)/2 + 1/2 \geq s(\mathbb{C}|L)$, with equality only when $s(\mathbb{C}|L) = 1$.

### 4.5. Examples

The split-merge framework (4) is flexible because the subcomponent similarity measures can be application specific. We give two examples.

**Entropy-based $S_H$** This example uses a normalized entropy of clustering. Let $s(\mathbb{C}|L) = 1 - H(\mathbb{C}_L)/\log|L|$ and $s(\mathbb{L}|C) = 1 - H(\mathbb{L}_C)/\log|C|$. Substituting them into (4) gives a similarity measure in the split-merge framework, and we call this measure $S_H$:

$$
\begin{aligned}
&S_H(\mathbb{L}, \mathbb{C}) \\
&= \sum_{L \in \mathbb{L}} \sum_{C \in \mathbb{C}} \frac{|L \cap C|}{n} \left(1 - \frac{H(\mathbb{C}_L)}{\log|L|}\right) \left(1 - \frac{H(\mathbb{L}_C)}{\log|C|}\right).
\end{aligned}
$$

We will compare this empirically to some existing similarity measures in Section 6.

**Mean-squared-error-based** The split-merge framework (4) can make use of the feature vectors of points or the distances between points when these are available. In contrast, previous external measures ignore such information (Coen et al., 2010). Many internal compactness and separation measures can be used as subcomponent measures (Liu et al., 2010). We give an example using the mean-squared-error to score a subcomponent. For a split subcomponent $\langle L, \mathbb{C} \rangle$, the mean-squared-error of $\mathbb{C}_L$ is $mse(\mathbb{C}_L) = \sum_{C \in \mathbb{C}_L} \sum_{i \in C} (f_i - \bar{f}_C)^2/n$, where $\bar{f}_C = \sum_{i \in C} f_i/|C|$ is the center of cluster $C$. The similarity of the split subcomponent is defined to be

$$MSE(\mathbb{C}|L) = \frac{mse(\mathbb{C}_L)}{mse(\{L\})} = \frac{\sum_{C \in \mathbb{C}_L} \sum_{i \in C} (f_i - \bar{f}_C)^2}{\sum_{i \in L} (f_i - \bar{f}_L)^2},$$

where $\bar{f}_L = \sum_{i \in L} f_i/|L|$ is the center of cluster $L$.

Using this subcomponent similarity measure, we can obtain a similarity measure between clusterings based on the split-merge framework (4). It is efficient to compute because the meet of two clusterings can be computed in $O(n)$ time when an appropriate data structure is used (Pantel & Lin, 2002). In contrast, the hybrid measure proposed by Coen et al. (2010) requires $O(n^{2.6})$ time in the average case and $O(n^3 \log n)$ time in the worst case.

## 5. Comparisons with Existing Measures

We study some properties of our split-merge similarity framework $S^*$ given by (4) in comparison with other measures. There are other properties which we explore in Section 3 of the supplementary material.

### 5.1. Conditional Normalization

To facilitate interpretation and comparison across different conditions (e.g., different data sets), the traditional normalization property focuses on the joint space of the two clusterings and requires that the range of a similarity measure should be normalized to a closed interval $[0, 1]$ (Wagner & Wagner, 2007; Vinh et al., 2010), where the lower bound zero should be achievable. However, this does not take into account the fact that one clustering is typically the true clustering. Since one fundamental goal of a similarity measure is to rank clusterings against a true clustering, the similarities with respect to the true clustering should also be normalized (Luo et al., 2009): given a true clustering $\mathbb{L}$, a similarity measure should be between zero and one with both extremes attainable. Luo et al. (2009) have found that some information theoretic measures did not satisfy this property, and they have proposed a *normalization* procedure using the extreme values attained by the original measures, which typically depends on $\mathbb{L}$ or $n$. Here, we propose *conditional normalization* based on a three-way partitioning of the set $\Omega$ of all possible clusterings of $n$ data points given a true clustering $\mathbb{L}$.

**Property 5.1 (Conditional Normalization).** *A similarity measure $S(\mathbb{L}, \mathbb{C})$ is conditionally normalized if, given $\mathbb{L}$,*

1. *$S(\mathbb{L}, \mathbb{C}) = 1$ if and only if $\mathbb{C} = \mathbb{L}$;*
2. *$S(\mathbb{L}, \mathbb{C}) = 0$ if and only if $\mathbb{C} \in \Omega_{\mathbb{L}}$; and*
3. *$S(\mathbb{L}, \mathbb{C}) \in (0, 1)$ otherwise, i.e., $\mathbb{C} \in \Lambda_{\mathbb{L}}$,*

*where $\{\{\mathbb{L}\}, \Omega_{\mathbb{L}}, \Lambda_{\mathbb{L}}\}$ partitions $\Omega$ such that $\Omega_{\mathbb{L}} = \varnothing$ if and only if $n = 1$, and $\Lambda_{\mathbb{L}} = \varnothing$ if and only if $n \leq 2$.*

We call $\mathbb{C} = \mathbb{L}$ *the best clustering*, and it is the only clustering can be scored one against $\mathbb{L}$. Each $\mathbb{C} \in \Omega_{\mathbb{L}}$ is called *a worst clustering*, and the similarity between it and $\mathbb{L}$ must be zero. With these, the extremes of $[0, 1]$ are realized. All other clusterings (i.e., those $\Lambda_{\mathbb{L}}$) have similarities in $(0, 1)$ with $\mathbb{L}$. Our definition is more stringent than that afforded by the procedure of (Luo et al., 2009) in two ways. First, only clustering $\mathbb{C} = \mathbb{L}$ can have the similarity one. This is to reflect that there is *only one* true clustering. Second, we demand that $\Lambda_{\mathbb{L}}$ be non-empty for $n \geq 3$. This gives a gradation of similarities from one to zero as a clustering deteriorates from the best clustering to a worst clustering; it reflects how far a clustering is from the best clustering and a worst clustering.

*Table 2.* Conditions where measures attain their lower bounds.

| Measure(s) | Lower Bound Condition |
|---|---|
| $R, A$ | $H(\mathbb{L}, \mathbb{C}) = \log n, H(\mathbb{L})H(\mathbb{C}) = 0$ |
| $N$ | $\mathbb{M} = \bot, N(\mathbb{L}, \mathbb{C}) = \frac{\lfloor\sqrt{n}\rfloor + \lceil\sqrt{n}\rceil}{2n}$ |
| $I$ | $H(\mathbb{L}, \mathbb{C}) \geq 0, H(\mathbb{L}) + H(\mathbb{C}) = H(\mathbb{L}, \mathbb{C})$ |
| $NMI$ | $H(\mathbb{L}, \mathbb{C}) > 0, H(\mathbb{L}) + H(\mathbb{C}) = H(\mathbb{L}, \mathbb{C})$ |
| $V$ | $H(\mathbb{L}, \mathbb{C}) = \log n, H(\mathbb{L}) + H(\mathbb{C}) = H(\mathbb{L}, \mathbb{C})$ |
| $K$ | $H(\mathbb{L}, \mathbb{C}) = 2\log k, H(\mathbb{L}) + H(\mathbb{C}) = H(\mathbb{L}, \mathbb{C})$ |
| $S^*$ | $\mathbb{M} = \bot, \mathbb{L} \cap \mathbb{C} = \varnothing$ |

A similarity measure is not normalized if its lower bound (resp. upper bound) is not zero (resp. one). An unnormalized measure is also not conditionally normalized. To determine whether a normalized similarity measure is conditionally normalized, we need to determine the subset $\Omega_{\mathbb{L}}$ for any $\mathbb{L}$. This requires us to determine the lower bound of a similarity measure as well as its *lower bound condition* that indicates what kind of clusterings are considered to be the worst. The lower bound conditions of the similarity measures are summarized in Table 2, and they are derived in Section 1 of the supplementary material. $\Omega_{\mathbb{L}}$ should not be empty for any $\mathbb{L} \in \Omega$ with $n \geq 1$, so a normalized similarity measure is not conditionally normalized if there exists a clustering $\mathbb{L}$ such that the lower bound condition has no solution. The following proposition gives the normalization properties of a selection of similarity measures.

**Proposition 5.1.** *N, A, and I are not normalized. R, NMI, V, and K are normalized but not conditionally normalized. $S^*$ is conditionally normalized.*

*Proof.* $N$ and $A$ are not normalized because their lower bounds are $(\lfloor\sqrt{n}\rfloor + \lceil\sqrt{n}\rceil)/2n$ and $1/n$, respectively. $I$ is not normalized because its upper bound is not always one. The lower bounds of other similarity measures are zero. $R$ can be zero only when $\mathbb{L}$ is $\top$ or $\bot$, so it is not conditionally normalized. The third category $\Lambda_{\top}$ of $\top$ is always empty because $NMI(\top, \mathbb{C}) = 0$ for any $\mathbb{C} \in \Omega$, so it is not conditionally normalized. When $k = \sqrt{n}$, $K$ becomes $V$. The lower bound condition of $V$ has no solution when $\mathbb{L} = \{\{1, 2\}, \{3\}\}$, so $V$ and $K$ are not conditionally normalized.

We prove that $S^*$ is conditionally normalized. First, it is clear that $S^*(\mathbb{L}, \mathbb{C}) = 1$ if and only if $\mathbb{L} = \mathbb{C}$. Second, its lower bound condition in Table 2 indicates that at least one worst-clustering can be constructed. Thirdly, the lower bound condition implies $\Omega_{\mathbb{L}} \neq \Omega \setminus \{\mathbb{L}\}$, and at least a clustering can be obtained such that its similarity score is in $(0, 1)$. For instance, if $\mathbb{L} \neq \bot$, then $\bot$ is such a clustering because $S^*(\mathbb{L}, \bot) \in (0, 1)$; and when $\mathbb{L} = \bot$ and $|\mathbb{L}| > 2$, such a clustering can be created by merging only two singletons. □

### 5.2. Join-weighted Decomposition and Consistency

We study whether some existing similarity measures satisfy Properties 4.1 and 4.4.

**Proposition 5.2.** *Only N, A, K, and $S^*$ are join-weighted decomposable. A and $S^*$ are subcomponent-consistent.*

*Proof.* That only $N$, $A$, $K$, and $S^*$ are join-weighted decomposable is shown in Table 1. Measures $N$ and $K$ are instances of $S'$ in (5), so they are not subcomponent-consistent. That $S^*(\{L\}, \mathbb{C}_L)$ is subcomponent-consistent is shown in Section 4.4. For $A$, we have $A(\{L\}, \mathbb{C}_L) = \max_{C \in \mathbb{C}_L} |C|/|L| = a(\mathbb{C}|L)$ and $A(\mathbb{L}_C, \{C\}) = a(\mathbb{L}|C)$, where $a$ is the relevant subcomponent measure. □

## 6. Experiments

We compare the conditional normalization and monotonically decreasing properties of various measures using the coreference resolution task. This task is to group noun phrases (data points) that refer into the same real-world entity (clusters) (Ng & Cardie, 2002). The $\phi_3$-CEAF measure (Luo, 2005) is frequently used to evaluate the performance of coreference algorithms, and it is the same as the classification accuracy. For a measure in our split-merge framework, we use the entropy-based $S_H$ in Section 4.5. Our experiments use a randomly selected document in the ACE-2005 English data set (Rahman & Ng, 2011).

We construct a series of clusterings from the best to a worst with respect to a given clustering. This series serves to evaluate similarity measures with respect to the following desideratum: a reasonable similarity score should decrease strictly from one to zero as the clustering "worsens" from the best to a worst. We use two operations to construct the series: (a) a *binary split operation* that splits the largest non-singleton cluster into two equal-sized clusters; and (b) a *binary merge operation* that either merges two true singletons into one cluster or merges a true singleton with a randomly selected cluster if there is only one singleton, where a true singleton is one that is in the true clustering. Given a true clustering, we first apply the binary split operation repeatedly to transform the true clustering to $\bot$.

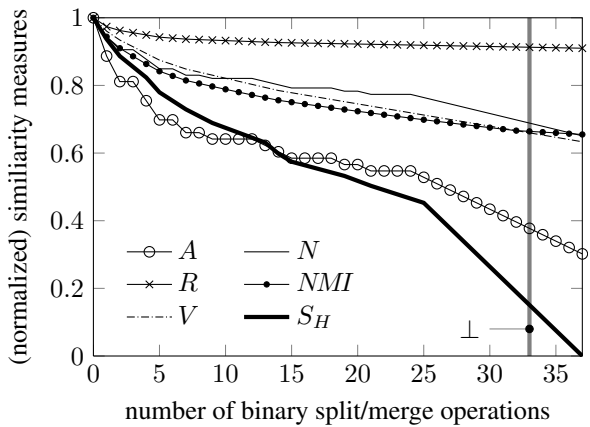

*Figure 3.* Measures on an ACE05 document for the coreference task as the clustering worsens from the best to a worst. The bottom clustering $\perp$ is obtained at the 33rd operation.

Then we apply the binary merge operation repeatedly to transform $\perp$ to a worst-clustering of the true clustering.

Figure 3 plots the (normalized) similarities as the number of operations increases. Although all the measures decreases as the generated clustering worsens, only $S_H$ decreases from one to zero. This is because, among these measures, only $S_H$ is conditionally normalized; see Proposition 5.1. In addition, the other measures are rather far from zero at the worst-clustering. The figure also shows that $S_H$ is strictly decreasing. This is also satisfied by $R$, $NMI$ and $V$ but not by set matching measures $A$ and $N$.

## 7. Conclusion

By modeling the intrinsic relation between two clusterings as a bipartite graph, we have proposed a split-merge framework that can be used to obtain similarity measures to compare clusterings on different data sets. In contrast with a representative selection of existing similarity measures, any measure obtained via the framework is conditionally normalized, join-weighted decomposable, and subcomponent-consistent. Conditional normalization is especially important because it allows comparing different clusterings of different data sets. In addition, our framework can also use feature vectors of data points or distances between data points.

## Acknowledgments

This work is supported by DSO grant DSOCL10021.